\documentclass[10pt,letterpaper]{article}

\usepackage{cogsci}
\usepackage{pslatex}
\usepackage[table, dvipsnames]{xcolor}
\usepackage{apacite}
\usepackage{pbox}
\usepackage{tabularx}
\usepackage{multirow}
\usepackage{graphicx}
\usepackage{comment}
\usepackage{placeins}
\usepackage{float}
\usepackage{subcaption}
\usepackage{amssymb}
\usepackage{amsmath}
\usepackage{mathtools}
\usepackage{booktabs}
\usepackage{qtree}
\usepackage{tikz-qtree}
\usepackage{tree-dvips}
\usepackage{gb4e}
\usepackage[normalem]{ulem}
\noautomath
\usepackage[colorinlistoftodos,prependcaption,textsize=scriptsize]{todonotes}

\newcolumntype{C}[1]{>{\centering\let\newline\\\arraybackslash\hspace{0pt}}m{#1}}

\title{Revisiting the poverty of the stimulus: hierarchical generalization without a hierarchical bias in recurrent neural networks}
 
\author{{\large \bf R.\ Thomas McCoy (tom.mccoy@jhu.edu)} \\
  Department of Cognitive Science, Johns Hopkins University
  \AND {\large \bf Robert Frank (bob.frank@yale.edu)} \\
  Department of Linguistics, Yale University \\
	\AND {\large \bf Tal Linzen (tal.linzen@jhu.edu)} \\
  Department of Cognitive Science, Johns Hopkins University }

\begin{document}

\maketitle

\begin{abstract}

Syntactic rules in natural language typically need to make reference to hierarchical sentence structure. 
However, the simple examples that language learners receive are often equally compatible with linear rules. Children consistently ignore these linear explanations and settle instead on the correct hierarchical one.
This fact has motivated the proposal that the learner's hypothesis space is constrained to include only hierarchical rules. We examine this proposal using recurrent neural networks (RNNs), which are not constrained in such a way. We simulate the acquisition of question formation, a hierarchical transformation, in a fragment of English. We find that some RNN architectures tend to learn the hierarchical rule, suggesting that hierarchical cues within the language, combined with the implicit architectural biases inherent in certain RNNs, may be sufficient to induce hierarchical generalizations. The likelihood of acquiring the hierarchical generalization increased when the language included an additional cue to hierarchy in the form of subject-verb agreement, underscoring the role of cues to hierarchy in the learner's input.

\end{abstract}

\section{Introduction}

Speakers of a language can generalize from finite linguistic experience to sentences they have never heard or produced before. Although there are many possible ways to generalize from a set of sentences, language learners consistently choose certain generalizations over others. In the syntactic domain, learners typically learn generalizations that appeal to hierarchical structures rather than linear order. An influential explanation for this fact is that learners never entertain hypotheses based on linear order: they are innately constrained to assume that syntactic rules are structure-sensitive \cite{chomsky1980}. 

To test whether a structure-sensitivity constraint is necessary to account for the generalizations that human language learners make, we use recurrent neural networks (RNNs), which are not equipped with such an explicit pre-existing hierarchical constraint.\footnote{In fact, RNNs are not just capable of using non-hierarchical structures but in fact appear to be biased in favor of linear structures over hierarchical ones \cite{christiansen1999}.} We simulate the acquisition of English \textbf{subject-auxiliary inversion}, the transformation that turns a declarative statement such as (\ref{ex:q1a}) into a question such as (\ref{ex:q1b}):

\begin{exe}
\ex \label{ex:q1} \begin{xlist}
\ex My walrus \textbf{can} giggle. \label{ex:q1a}
\ex \textbf{Can} my walrus giggle? \label{ex:q1b}
\end{xlist}
\end{exe}

\noindent At least two rules could generate (\ref{ex:q1b}) from (\ref{ex:q1a}):
\vskip0.5\baselineskip

\begin{minipage}[b]{0.22\textwidth}
\raggedright\includegraphics[trim={1cm 0 0 0 0}, clip, width=\textwidth]{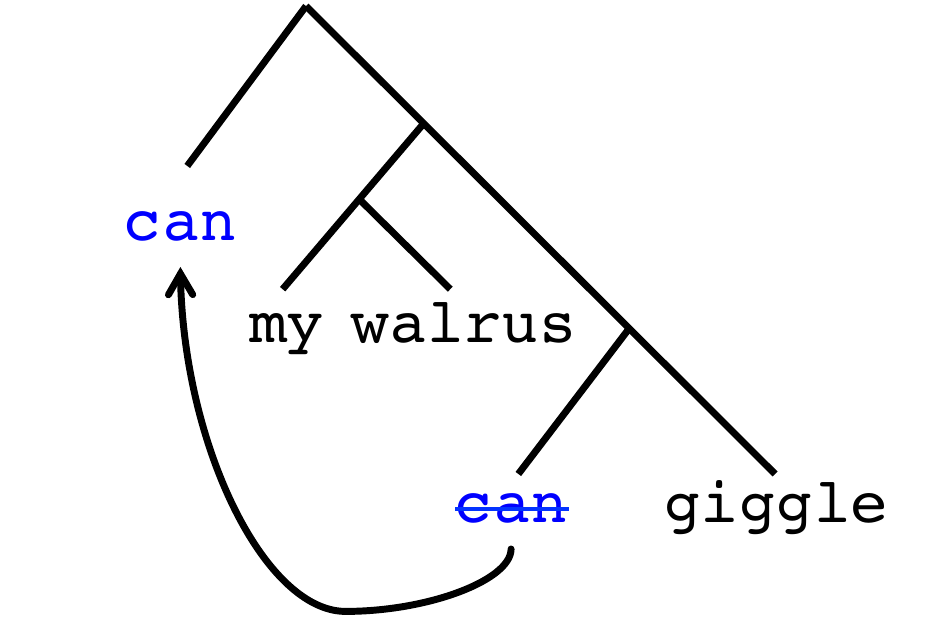}
\raggedright\textbf{Hierarchical rule:} Move the \underline{main verb's} auxiliary to the front of the sentence. \label{ex:hypa}
\end{minipage}\hspace{0.02\textwidth}%
\begin{minipage}[b]{0.22\textwidth}
\raggedright\includegraphics[trim={0.7cm 0 0 0 0}, clip, width=\textwidth]{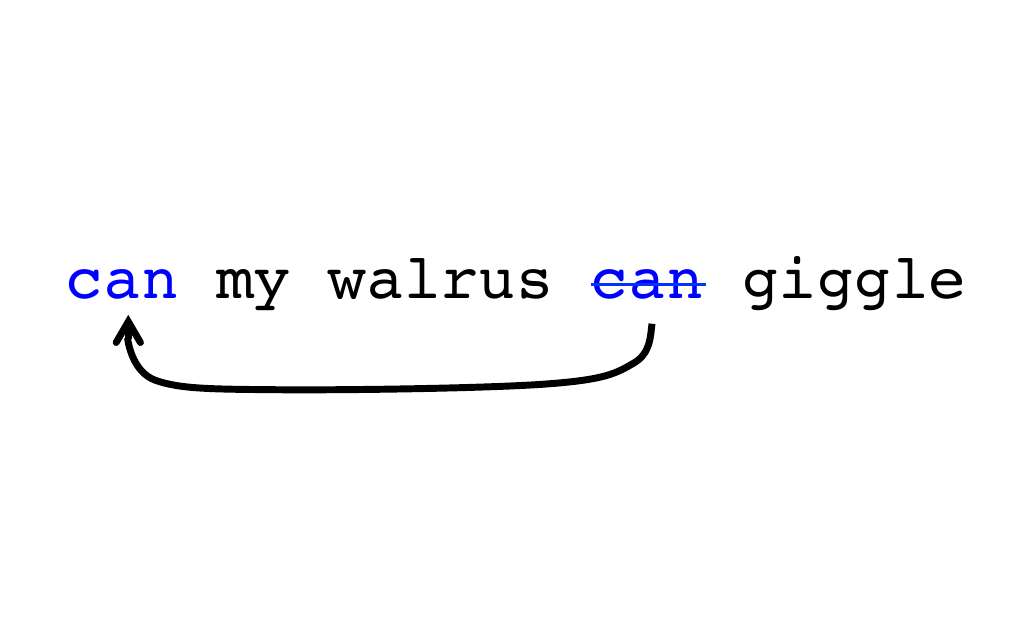}
\textbf{Linear rule:} Move the \underline{linearly first} auxiliary to the front of the sentence.\label{ex:hypb}
\end{minipage}
\vskip 1\baselineskip

\noindent While both rules account for common cases such as (\ref{ex:q1}), they make different predictions for complex sentences such as~(\ref{ex:q2}):

\begin{table*}[!h]
\begin{center} 
\begin{tabular}{lp{7cm}lp{5cm}}
\fbox{\textcolor{white}{T}} & Training set, test set &
\fcolorbox{black}{gray!25}{\textcolor{gray!25}{T}}& Generalization set 
\end{tabular}

\vskip 0.03in

\begin{tabularx} {\textwidth} {p{1cm} | X | X} \hline
& \centering IDENT & \centering QUEST\arraybackslash \\ \hline
No RC & \textit{Input:} the newt can confuse my yak by the zebra . \newline \textit{Output:} the newt can confuse my yak by the zebra . & \textit{Input:} the newt can confuse my yak by the zebra . \newline \textit{Output:} can the newt confuse my yak by the zebra ? \\  \hline
RC on object & \textit{Input:} the newt can confuse my yak who will sleep . \newline \textit{Output:} the newt can confuse my yak who will sleep . &  \textit{Input:} the newt can confuse my yak who will sleep . \newline \textit{Output:} can the newt confuse my yak who will sleep ? \\ \hline
RC on subject & \textit{Input:} the newt who will sleep can confuse my yak . \newline \textit{Output:} the newt who will sleep can confuse my yak . & \cellcolor{gray!25}{\textit{Input:} the newt who will sleep can confuse my yak . \newline \textit{Output:} can the newt who will sleep confuse my yak ?}\\ \hline
\end{tabularx}
\end{center}
    \caption{Examples for each combination of a sentence type and a task. RC stands for ``relative clause."}
\label{table:typetask}

\end{table*}

\begin{exe}
\ex My walrus that \textbf{will} eat \textbf{can} giggle. \label{ex:q2}
\end{exe}

\noindent Specifically, the hierarchical rule predicts the correct question (\ref{ex:q2a}), while the linear rule predicts the incorrect question (\ref{ex:q2b}):

\begin{exe}
\ex \begin{xlist}
\ex \textbf{Can} my walrus that \textbf{will} eat \underline{\hspace{0.5cm}} giggle? \label{ex:q2a}
\ex \makebox[0pt][r]{$^*$}\textbf{Will} my walrus that \underline{\hspace{0.5cm}} eat \textbf{can} giggle? \label{ex:q2b}
\end{xlist}
\end{exe}

\noindent Although such examples disambiguate the two hypotheses, \citeA{chomsky1971} argues that they are highly infrequent, and thus children may never encounter them. Without these critical examples, according to Chomsky, children can only acquire the hierarchical rule by drawing on an innate constraint stipulating that syntactic rules must appeal to hierarchy. 

This argument, known as the \textbf{argument from the poverty of the stimulus} \cite{chomsky1980}, has been challenged in a number of ways. Some have disputed the assumption that children never encounter critical cases such as (\ref{ex:q2a}) \cite{pullum2002}. Others have questioned the assumption that an explicit hierarchical constraint is necessary for hierarchical generalization. One such approach has been to argue that the hierarchical rule can fall out of weaker or non-syntactic structural biases. For example, \citeA{perfors2011} showed that a learner whose task is to choose between an innately available hierarchical representation and an innately available linear representation will choose the hierarchical one; and \citeA{fitz2017} argued that the hierarchical structure of questions is rooted in innately available structured semantic representations.

A second approach has dispensed with pre-existing structural representations altogether. \citeA{lewis2001} argued that an RNN trained to predict the next word can learn which questions are well formed, but this conclusion was convincingly called into question by \citeA{kam2008}. The most immediate precursor to our work is \citeA{frank2007}. Like \citeauthor{lewis2001}, they used RNNs, but instead of modeling the well-formedness of the question alone, they followed the traditional framework of transformational grammar in modeling the generation of a question from a declarative sentence.\footnote{This is a simplification---a more psychologically plausible assumption would be that questions are generated from a semantic representation shared with the declarative sentence \cite{fitz2017}.} Their results were difficult to interpret because the network's generalization behavior depended heavily on the identity of the auxiliaries in the input sentence, and neither the linear hypothesis nor the hierarchical hypothesis predict such lexically dependent behavior. We significantly expand on their experiments, taking advantage of recent technological and architectural advances in RNNs that have shown promise in the acquisition of syntax \cite{linzen2016assessing}.

To anticipate our results, of the six RNN architectures we explored, one of the architectures consistently learned a hierarchical generalization for question formation. This suggests that a learner's preference for hierarchy may arise from the hierarchical properties of the input, coupled with biases implicit in the network's computational architecture and learning procedure, without the need for pre-existing hierarchical constraints in the learner. We provide further evidence for the role of the hierarchical properties of the input by showing that adding syntactic agreement to the input increased the probability that a network would make hierarchical generalizations.

\begin{figure*}
\begin{center} 
\includegraphics[width=\textwidth]{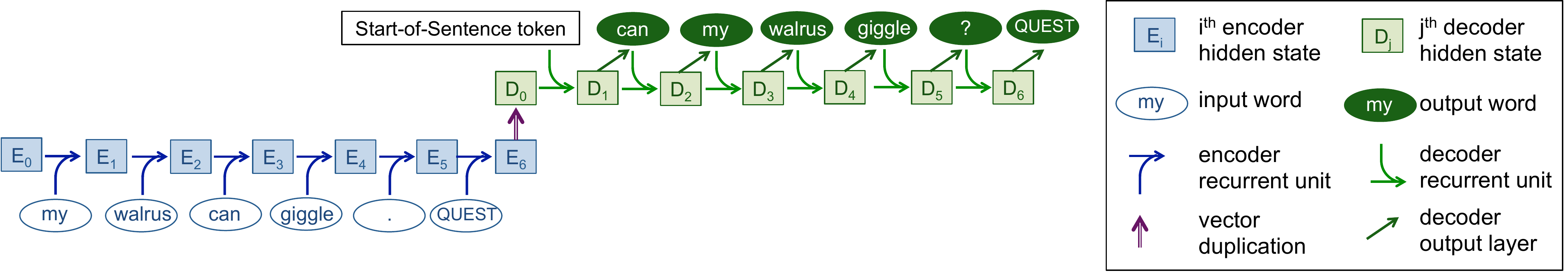}
\end{center}
\caption{Basic sequence-to-sequence neural network without attention.} \label{fig:seq}
\end{figure*}

\section{Experimental setup}

\subsection{Languages}

The networks were trained on two fragments of English, each consisting of a subset
of all possible declarative sentences and questions.\footnote{The vocabulary of the fragments consisted of 66 words. The full context-free grammar characterizing the fragments, along with statistics about the generated sentences, can be found in the supplementary materials.} We refer to the first fragment as the \textbf{no-agreement language}. Examples of declarative sentences in this language are given in (\ref{ex:noagr}):

\begin{exe}
\ex 
\label{ex:noagr}
\begin{xlist}
\ex the walrus can giggle .
\ex the yak could amuse your quails by my raven .
\ex the walruses that the newt will confuse can high\_five your peacocks .
\end{xlist}
\end{exe}

Each noun phrase in the language had at most one modifier, either a relative clause or a prepositional phrase. Relative clauses were never embedded inside other relative clauses.
Every verb was associated with one of the auxiliary verbs \textit{can}, \textit{could}, \textit{will}, and \textit{would}. Since such modals do not show agreement, any noun, whether singular or plural, was allowed to appear with any auxiliary. 

The second fragment, the \textbf{agreement language}, was identical to the no-agreement language, except that the auxiliaries in this language were \textit{do}, \textit{don't}, \textit{does}, and \textit{doesn't}. Subjects in this language agreed with the auxiliaries of their verbs: singular subjects appeared with \textit{does} or \textit{doesn't}, while plural subjects appeared with \textit{do} or \textit{don't}. Examples of declarative sentences in the agreement language are given in (\ref{ex:agr}):

\begin{exe}
\ex 
\label{ex:agr}
\begin{xlist}
\ex the walrus does giggle .
\ex the yak doesn't amuse your quails by my raven .
\ex the walruses that the newt does confuse do high\_five your peacocks .\label{ex:longdist}
\end{xlist}
\end{exe}

Both languages reused structural units; for example, the same prepositional phrases could modify both subject and object nouns. Such shared structure served as a possible cue to hierarchy because it is more efficiently represented in a hierarchical grammar than a linear one. Subject-verb agreement in the agreement language provided an additional cue to hierarchy; in (\ref{ex:longdist}), for example, \textit{do} agrees with its hierarchically-determined plural subject of \textit{walruses} even though the singular noun \textit{newt} is linearly closer to it. We therefore predict that hierarchical generalizations will be more likely with the agreement language than the no-agreement language.

\subsection{Tasks}

The networks were trained to perform two tasks: identity (returning the input sentence unchanged) and question formation. The task to be performed was indicated by a token at the end of the sentence---either \textit{IDENT} for identity or \textit{QUEST} for question formation. \textit{IDENT} and \textit{QUEST} served as end-of-sequence tokens in both the input and output.

Table~\ref{table:typetask} provides examples of these tasks on each of the three types of sentences in the languages: sentences without relative clauses, sentences with a relative clause on the  object, and sentences with a relative clause on the subject. During training we withheld the question formation task for sentences with a relative clause on the subject (the shaded cell in Table~\ref{table:typetask}); these are the only cases that directly disambiguate the linear and hierarchical hypotheses.  The identity task was included in the training setup to familiarize the networks with the critical sentence type withheld from the question task; without such exposure, the networks could be justified in concluding that subjects cannot be modified by relative clauses, making it difficult to test such sentences.

\subsection{Evaluation}

We used two sets of sentences for evaluation, a test set and a generalization set. The test set consisted of novel sentences from the five non-withheld cases in Table \ref{table:typetask}. It was used to assess how well a network had learned the patterns in its training set. The generalization set consisted of sentences from the withheld case (the question formation task for sentences with relative clauses on their subjects). This set was used to assess how the networks generalized to sentence types from which they had not formed questions during training. The test and generalization set both contained 10,000 unique sentences and the training set contained 120,000 unique sentences.

\subsection{Architectures}

Here we give a very brief bird's-eye view of our architectures. For a more precise description, including our hyperparameter values, see the supplementary materials.

For all experiments we used the sequence-to-sequence model \cite{botvinick2006,sutskever2014} illustrated in Figure \ref{fig:seq}. This network has two subcomponents called the \textbf{encoder} and the \textbf{decoder}, both of which are RNNs. The encoder processes the input sentence one word at a time to create a single vector representing the entire input sentence. The decoder then receives this vector (called the \textbf{encoding}) and, based on it, outputs one word at a time until it generates a special end-of-sequence token. 

The encoder and decoder each possess a component called a \textbf{recurrent unit} which governs how information flows from one time step to the next. We tested three types of recurrent units: a simple recurrent network (SRN) \cite{elman1990}, a gated recurrent unit (GRU) \cite{cho2014}, and long short-term memory (LSTM) \cite{hochreiter1997}. For each type of recurrent unit, we experimented with adding \textbf{attention} to the decoder \cite{bahdanau2015}; attention is a mechanism which gives the decoder access to intermediate steps of the encoding process. For each pair of an architecture and a language, we trained 100 networks with different random initializations, for a total of 1200 networks.

\section{Results}

\subsection{Test set}

For the test set, all six architectures except the vanilla SRN (i.e., the SRN without attention) produced over 94\% of the output sentences exactly correctly (accuracy was averaged across 100 trained networks for each architecture). The highest accuracy was 99.9\% for the LSTM without attention. Using a more lenient evaluation criterion whereby the network was not penalized for replacing a word with another word of the same part of speech, the accuracy of the SRN without attention increased from 0.1\% to 81\%,
suggesting that its main source of error was a tendency to replace words with other words of the same lexical category. This tendency is a known deficiency of SRNs \cite{frank2007} and does not bear on our main concern of the networks' syntactic representations. Setting aside these lexical concerns, then, we conclude that all architectures were able to learn the language.

\subsection{Generalization set}

On the generalization set, the networks were rarely able to correctly produce the full question -- only about 13\% of the questions were exactly correct in the best-performing architecture (LSTM with attention). However, getting the output exactly correct is a demanding metric; the full-question accuracy can be affected by a number of errors that are not directly related to the research question of whether the network preferred a linear or hierarchical rule. Such errors include repeating or omitting words or confusing similar words. To abstract away from such extraneous errors, for the generalization set we focus on accuracy at the first word of the output. Because all examples in the generalization set involve question formation, this word is always the auxiliary that is moved to form the question, and the identity of this auxiliary is enough to differentiate the hypotheses. For example, if the input is \textit{my yak who the seal can amuse will giggle . QUEST}, a hierarchically-generalizing network would choose \textit{will} as the first word of the output, while a linearly-generalizing network would choose \textit{can}. This analysis only disambiguates the hypotheses if the two possible auxiliaries are different, so we only considered sentences where that was the case. For the agreement language, we made the further stipulation that both auxiliaries must agree with the subject so that the correct auxiliary could not be determined based on agreement alone.

Figure \ref{fig:firstacc} gives the accuracies on this metric across the six architectures for the two different languages (individual points represent different initializations). We draw three conclusions from this figure:

\begin{figure}[t]
\centering
\includegraphics[width=0.45\textwidth]{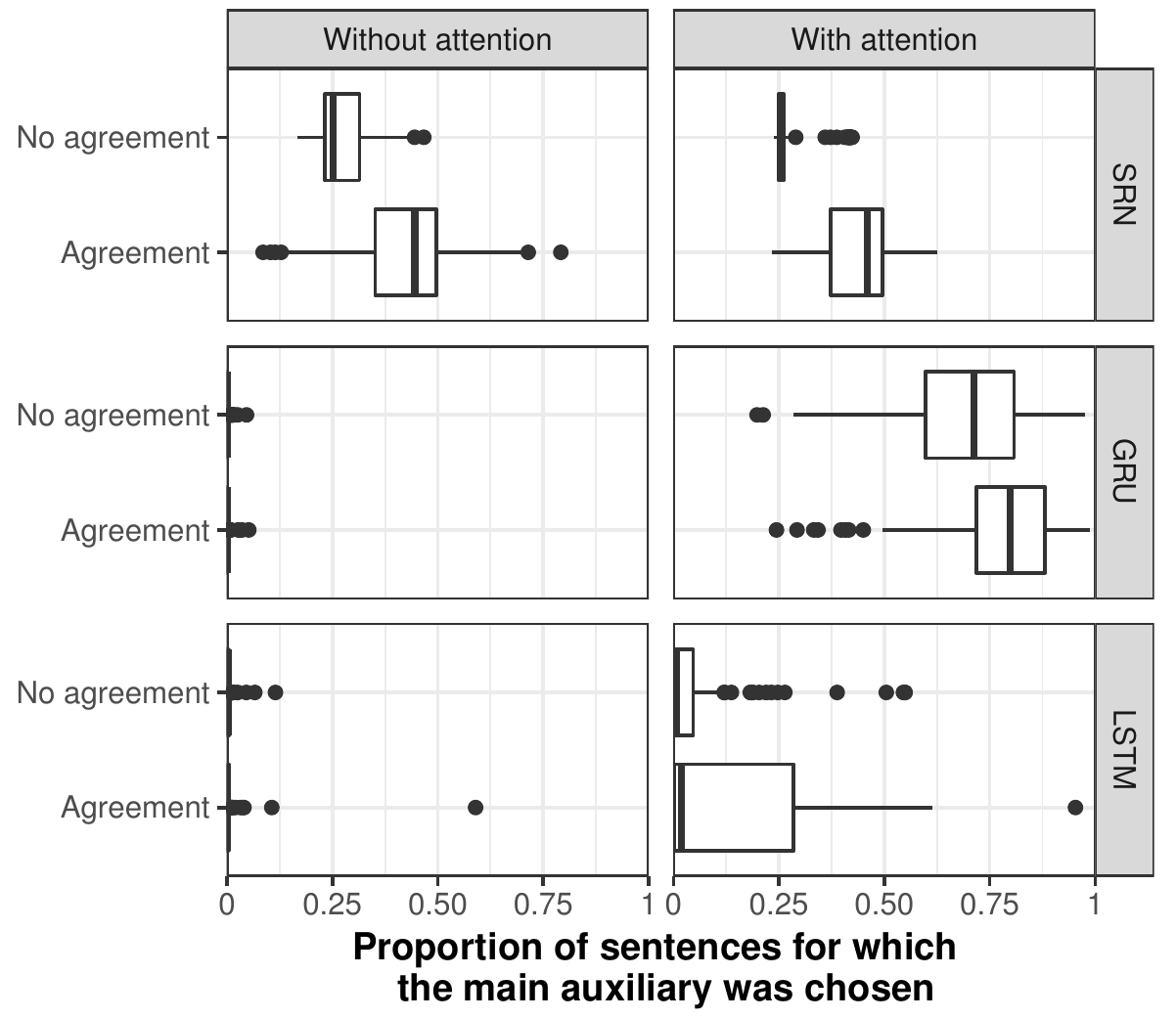}
    \caption{Accuracy of auxiliary prediction for questions of the withheld type (with a relative clause on the subject). 
} \label{fig:firstacc}
\end{figure}

\noindent
\textbf{1. Agreement leads to more robust hierarchical generalization:}
All six architectures were significantly more likely ($p < 0.01$) to choose the main auxiliary when trained on the agreement language than the no-agreement language. In other words, adding hierarchical cues to the input increased the chance of learning the hierarchical generalization.

\noindent
\textbf{2. Initialization matters:}
For each architecture, accuracy often varied considerably across random initializations. This fact suggests that the architectural bias is not strong enough to reliably lead the networks to settle on the hierarchical generalization, even in GRUs with attention. From a methodological perspective, this observation highlights the importance of examining many initializations of the network before drawing qualitative conclusions about an architecture (in a particularly striking example, though the accuracy of most LSTMs with attention was low, there was one with near-perfect accuracy).

\noindent
\textbf{3. Different architectures perform qualitatively differently:}
Of the six architectures, only the GRU with attention showed a strong preference for choosing the main auxiliary instead of the linearly first auxiliary. By contrast, the vanilla GRU chose the first auxiliary nearly 100\% of the time. In this case, then, attention made a qualitative difference for the generalization that was acquired. By contrast, for both LSTM architectures, most random initializations led to networks that  chose the first auxiliary nearly 100\% of the time. Both SRN architectures showed little preference for either the main auxiliary or the linearly first auxiliary; in fact the SRNs often chose an auxiliary that was not even in the input sentence, whereas the GRUs and LSTMs almost always chose one of the auxiliaries in the input. In the next section, we take some preliminary steps toward exploring why the  architectures behaved in qualitatively different ways.

\subsection{Analysis of sentence encodings}

A plausible hypothesis about the differences between networks is that linearly-generalizing networks used representations that contained linearly-relevant information whereas hierarchically-generalizing networks used representations that contained hierarchically-relevant information. To test this hypothesis, we analyzed the final hidden state of the encoder ($E_6$ in Figure \ref{fig:seq}), which we will refer to as the encoding of the sentence. In architectures without attention, this is the only information that the decoder has about the sentence; architectures with attention can use the intermediate encodings of sentence prefixes as well. We analyze the amount of information that these encodings contain about three properties of the input sentence: its main auxiliary, its fourth word, and the head noun of the subject (which, in the simple languages we used, was always the sentence's second word). Examples are shown in Table \ref{tab:linex}.

\begin{table*}
\centering
\begin{tabular} {p{0.3\textwidth} p{0.3\textwidth} p{0.3\textwidth}} 
\multicolumn{1}{c}{Main auxiliary} & \multicolumn{1}{c}{Fourth word} & \multicolumn{1}{c}{Subject noun} \\ \hline
my unicorns \textbf{would} laugh . & my unicorns would \textbf{laugh} . & my \textbf{unicorns} would laugh . \\
my quail with her yak \textbf{will} read . & my quail with \textbf{her} yak will read . & my \textbf{quail} with her yak will read . \\
his newt who can giggle \textbf{could} swim . & his newt who \textbf{can} giggle could swim . & his \textbf{newt} who can giggle could swim . \\ \hline
\end{tabular}
\caption{Examples of the entities identified by the linear classifiers.} \label{tab:linex}
\end{table*}

\noindent
\textbf{Main auxiliary:} 
The main auxiliary of a sentence can appear in many different linear positions but has a consistent hierarchical position. Therefore, a network whose encodings can be used to identify sentences' main auxiliaries must contain some hierarchical information.

\noindent
\textbf{Fourth word:}
The fourth word of a sentence has a consistent role in a linear representation but not in a hierarchical one: the fourth word could be the main verb, the determiner on a prepositional object, or the auxiliary verb inside a subject relative clause. Therefore, a network whose encodings can be used to identify each sentence's fourth word must contain some information about linear order.

\noindent
\textbf{Subject noun/second word:}
The head noun of the subject is always the second word of the sentence in our languages. Thus, this word can be reliably identified either from a linear representation (as the second word) or from a hierarchical representation (as the subject noun).

\noindent
\textbf{Analysis:}
For each trained network, we trained three linear classifiers, one for each of these three properties of the sentence. Each classifier was trained to predict the word that filled the relevant role---main auxiliary, fourth word or subject noun/second word---from the final hidden state of the encoder. Each classifier's output layer had a dimensionality equal to the number of possible classes for that classifier's task: 4 for the main auxiliary, 28 for the fourth word, or 26 for the subject noun. 
The classifiers were trained on a training set and tested on a withheld test set (see the supplementary materials for details). Figure~\ref{fig:linclass} shows the classification results on the test set. 

\begin{figure}
\centering
\includegraphics[width=0.48\textwidth]{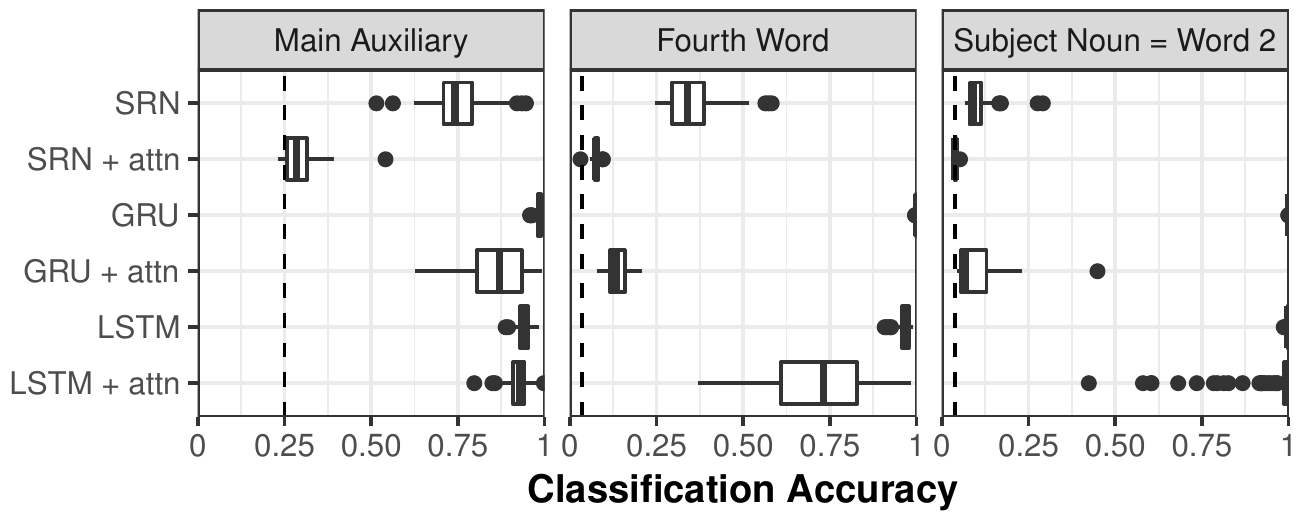}
\caption{Linear classifier results. Dotted lines indicate chance performance.} \label{fig:linclass}
\end{figure}

Classifiers trained to predict the main auxiliary from the encodings produced by the SRNs with attention performed only slightly better than chance; this might explain why the SRNs with attention generalized poorly to the withheld sentence type in the question formation task. Similar classifiers trained on encodings from the other architectures did well at this task. Since the identity of the main auxiliary is the only information required to perform well on our evaluation of the networks' performance on the generalization set based on the first word produced, these results suggest that the differences in performance stem not from inability to identify the main auxiliary but rather from a misinterpretation of the task as requiring fronting of the linearly first auxiliary.

We now consider the fourth word and subject noun classifiers. The classifiers trained on the encodings from both types of LSTMs as well as the GRUs without attention performed well at both tasks. Crucially, the classifiers trained on the encodings from the GRU with attention did poorly on these tasks. Recall that the main auxiliary could be successfully decoded from the encodings of this architecture. The GRU with attention therefore appears to use its encoding only for information that could not be straightforwardly obtained from linear order, such as the main auxiliary, rather than information that could be obtained from linear order even if, like the subject head noun, that information was hierarchically relevant. On the other hand, the fact that the GRU without attention and both LSTM architectures performed very well at all three tasks suggests that they used their encodings for both linear and hierarchical information.
Thus, perhaps the better generalization ability of the GRU with attention arises not from a better ability to encode relevant hierarchical information---all  four LSTM and GRU
architectures have that ability---but rather from an ability to ignore linear information \cite{frank2007}. 

\subsection{Comparing RNN Mistakes with Human Mistakes}

We now return to the full questions produced by our networks and compare the networks' errors to the types of errors that humans make when acquiring English \cite{crain1987}. We restrict ourselves to the GRU with attention networks as those were the networks that generally produced the correct auxiliary (see Figure \ref{fig:firstacc}). 

Subject-auxiliary inversion can be decomposed into two subtasks: placing an auxiliary at the start of the sentence and deleting an auxiliary within the sentence. Only 65\% of the outputs that the 100 networks collectively produced could be interpreted as having been formed by inserting an auxiliary before the sentence and deleting zero or one of the auxiliaries in the sentence. 
Table \ref{tab:candy} breaks down those results based on which auxiliary was preposed and which (if any) was deleted.\footnote{See the supplementary materials for examples of the remaining 35\% of outputs.}  

\begin{table}
\centering
\begin{tabular} { | c | r | r | r |}  \hline
& Prepose 1$^{\textrm{st}}$ & Prepose 2$^{\textrm{nd}}$ & Prepose other \\ \hline
Delete 1$^{\textrm{st}}$ & 7\% & 24\% & 4\% \\ \hline
Delete 2$^{\textrm{nd}}$ & 0\% & 3\% & 0\% \\ \hline
Delete none & 4\% & 21\% & 2\% \\ \hline
\end{tabular}
    \caption{Analysis of output question types based on which auxiliary has been deleted (if any) and which auxiliary has been placed at the start of the sentence. Each number is the percent of GRU + attention outputs across all 100 random initializations that fit that category (the total sums to 65\% because only 65\% of the questions produced by the networks could be analyzed in that way). \textit{1$^{\textrm{st}}$} and \textit{2$^{\textrm{nd}}$} refer to the first and second auxiliaries in the input. 
} \label{tab:candy}
\end{table}

Two error types are by far the most common. In the first type, the network preposed the second auxiliary but did not delete either of the auxiliaries ({\em could his newt who can giggle could swim} from \textit{his newt who can giggle could swim}).  This error type is common among English-learning children  \cite{crain1987}  and is compatible with hierarchical generalization. In the other frequent error type, the network deleted the first auxiliary and preposed the second; for example, it might generate {\em could his newt who giggle could swim} from \textit{his newt who can giggle could swim}. Such errors were never observed by \citeA{crain1987} and are incompatible with a hierarchical generalization. In other words, though the networks' common error types overlapped with the common error types for humans, the networks also frequently made some mistakes that humans never would. 

\section{Conclusions and Future Work}

Learners of English acquire the correct hierarchical rule for forming questions even though there are few to no examples in their input that explicitly distinguish this rule from the linear one. This fact has been taken to suggest that learners must be innately constrained to consider only hierarchical syntactic rules. We have investigated whether a learner without such a constraint can learn the hierarchical generalization without the critical disambiguating examples. Based on the behavior of one of the architectures we examined (GRU with attention), the answer to this question appears to be yes.  The hierarchical behavior of this non-hierarchically-constrained architecture plausibly arose from the influence of hierarchical cues in the input, a conclusion supported by the fact that the additional hierarchical cue of agreement increased the likelihood that a network would induce hierarchical generalizations.

Our argument has focused on a strong version of the poverty of the stimulus argument which claims that language learners require a hierarchical \textit{constraint}. However, there remains a milder version which only claims that a hierarchical \textit{bias} is necessary. This version of the argument is difficult to assess using RNNs because, while RNNs must possess some biases \cite{mitchell1980,marcus2018}, the nature of these biases---which likely arise both from the network architecture and from the learning algorithm---is currently poorly understood. However, given the linear way in which they process inputs, it is plausible that all six architectures we used had a bias toward linear order but that the GRU with attention was the only one that overcame this linear bias sufficiently to  generalize hierarchically. It is not clear why it was the only architecture to do so; we intend to examine the differences in behavior between the recurrent units in future work.

Two caveats are in order. First, our results only cover restricted fragments of English and may not generalize to the linguistic input that human language learners encounter. In future work, we will replace our artificial languages with a corpus of child-directed speech. Second, even if our findings do generalize to realistic language, we would only be able to conclude that it is \textit{possible} to solve the task without a hierarchical constraint; humans certainly could have such an innate constraint despite it being unnecessary for this particular task.

\section{Acknowledgments}

Our experiments were conducted using the resources of the Maryland Advanced Research Computing Center (MARCC). We thank Joe Pater, Paul Smolensky, and the JHU Computational Psycholinguistics group for helpful comments.

\bibliographystyle{apacite}

\setlength{\bibleftmargin}{.125in}
\setlength{\bibindent}{-\bibleftmargin}

\bibliography{CogSci_Template}

\begin{thebibliography}{}

\bibitem [\protect \citeauthoryear {%
Bahdanau%
, Cho%
\BCBL {}\ \BBA {} Bengio%
}{%
Bahdanau%
\ \protect \BOthers {.}}{%
{\protect \APACyear {2015}}%
}]{%
bahdanau2015}
\APACinsertmetastar {%
bahdanau2015}%
\begin{APACrefauthors}%
Bahdanau, D.%
, Cho, K.%
\BCBL {}\ \BBA {} Bengio, Y.%
\end{APACrefauthors}%
\unskip\
\newblock
\APACrefYearMonthDay{2015}{}{}.
\newblock
{\BBOQ}\APACrefatitle {Neural machine translation by jointly learning to align
  and translate} {Neural machine translation by jointly learning to align and
  translate}.{\BBCQ}
\newblock
\BIn{} \APACrefbtitle {Proceedings of {ICLR}.} {Proceedings of {ICLR}.}
\PrintBackRefs{\CurrentBib}

\bibitem [\protect \citeauthoryear {%
Botvinick%
\ \BBA {} Plaut%
}{%
Botvinick%
\ \BBA {} Plaut%
}{%
{\protect \APACyear {2006}}%
}]{%
botvinick2006}
\APACinsertmetastar {%
botvinick2006}%
\begin{APACrefauthors}%
Botvinick, M\BPBI M.%
\BCBT {}\ \BBA {} Plaut, D\BPBI C.%
\end{APACrefauthors}%
\unskip\
\newblock
\APACrefYearMonthDay{2006}{}{}.
\newblock
{\BBOQ}\APACrefatitle {Short-term memory for serial order: A recurrent neural
  network model} {Short-term memory for serial order: A recurrent neural
  network model}.{\BBCQ}
\newblock
\APACjournalVolNumPages{Psychological Review}{113}{2}{201--233}.
\PrintBackRefs{\CurrentBib}

\bibitem [\protect \citeauthoryear {%
Cho%
\ \protect \BOthers {.}}{%
Cho%
\ \protect \BOthers {.}}{%
{\protect \APACyear {2014}}%
}]{%
cho2014}
\APACinsertmetastar {%
cho2014}%
\begin{APACrefauthors}%
Cho, K.%
, {Van Merriënboer}, B.%
, Gulcehre, C.%
, Bahdanau, D.%
, Bougares, F.%
, Schwenk, H.%
\BCBL {}\ \BBA {} Bengio, Y.%
\end{APACrefauthors}%
\unskip\
\newblock
\APACrefYearMonthDay{2014}{}{}.
\newblock
{\BBOQ}\APACrefatitle {Learning phrase representations using {RNN}
  encoder-decoder for statistical machine translation} {Learning phrase
  representations using {RNN} encoder-decoder for statistical machine
  translation}.{\BBCQ}
\newblock
\BIn{} \APACrefbtitle {Proceedings of {EMNLP}.} {Proceedings of {EMNLP}.}
\PrintBackRefs{\CurrentBib}

\bibitem [\protect \citeauthoryear {%
Chomsky%
}{%
Chomsky%
}{%
{\protect \APACyear {1971}}%
}]{%
chomsky1971}
\APACinsertmetastar {%
chomsky1971}%
\begin{APACrefauthors}%
Chomsky, N.%
\end{APACrefauthors}%
\unskip\
\newblock
\APACrefYear{1971}.
\newblock
\APACrefbtitle {Problems of Knowledge and Freedom} {Problems of knowledge and
  freedom}.
\newblock
\APACaddressPublisher{New York}{Pantheon}.
\PrintBackRefs{\CurrentBib}

\bibitem [\protect \citeauthoryear {%
Chomsky%
}{%
Chomsky%
}{%
{\protect \APACyear {1980}}%
}]{%
chomsky1980}
\APACinsertmetastar {%
chomsky1980}%
\begin{APACrefauthors}%
Chomsky, N.%
\end{APACrefauthors}%
\unskip\
\newblock
\APACrefYearMonthDay{1980}{}{}.
\newblock
{\BBOQ}\APACrefatitle {Rules and representations} {Rules and
  representations}.{\BBCQ}
\newblock
\APACjournalVolNumPages{Behavioral and Brain Sciences}{3}{1}{1--15}.
\PrintBackRefs{\CurrentBib}

\bibitem [\protect \citeauthoryear {%
Christiansen%
\ \BBA {} Chater%
}{%
Christiansen%
\ \BBA {} Chater%
}{%
{\protect \APACyear {1999}}%
}]{%
christiansen1999}
\APACinsertmetastar {%
christiansen1999}%
\begin{APACrefauthors}%
Christiansen, M\BPBI H.%
\BCBT {}\ \BBA {} Chater, N.%
\end{APACrefauthors}%
\unskip\
\newblock
\APACrefYearMonthDay{1999}{}{}.
\newblock
{\BBOQ}\APACrefatitle {Toward a connectionist model of recursion in human
  linguistic performance} {Toward a connectionist model of recursion in human
  linguistic performance}.{\BBCQ}
\newblock
\APACjournalVolNumPages{Cognitive Science}{23}{2}{157--205}.
\PrintBackRefs{\CurrentBib}

\bibitem [\protect \citeauthoryear {%
Crain%
\ \BBA {} Nakayama%
}{%
Crain%
\ \BBA {} Nakayama%
}{%
{\protect \APACyear {1987}}%
}]{%
crain1987}
\APACinsertmetastar {%
crain1987}%
\begin{APACrefauthors}%
Crain, S.%
\BCBT {}\ \BBA {} Nakayama, M.%
\end{APACrefauthors}%
\unskip\
\newblock
\APACrefYearMonthDay{1987}{}{}.
\newblock
{\BBOQ}\APACrefatitle {Structure dependence in grammar formation} {Structure
  dependence in grammar formation}.{\BBCQ}
\newblock
\APACjournalVolNumPages{Language}{}{}{522--543}.
\PrintBackRefs{\CurrentBib}

\bibitem [\protect \citeauthoryear {%
Elman%
}{%
Elman%
}{%
{\protect \APACyear {1990}}%
}]{%
elman1990}
\APACinsertmetastar {%
elman1990}%
\begin{APACrefauthors}%
Elman, J\BPBI L.%
\end{APACrefauthors}%
\unskip\
\newblock
\APACrefYearMonthDay{1990}{}{}.
\newblock
{\BBOQ}\APACrefatitle {Finding structure in time} {Finding structure in
  time}.{\BBCQ}
\newblock
\APACjournalVolNumPages{Cognitive Science}{14}{2}{179--211}.
\PrintBackRefs{\CurrentBib}

\bibitem [\protect \citeauthoryear {%
Fitz%
\ \BBA {} Chang%
}{%
Fitz%
\ \BBA {} Chang%
}{%
{\protect \APACyear {2017}}%
}]{%
fitz2017}
\APACinsertmetastar {%
fitz2017}%
\begin{APACrefauthors}%
Fitz, H.%
\BCBT {}\ \BBA {} Chang, F.%
\end{APACrefauthors}%
\unskip\
\newblock
\APACrefYearMonthDay{2017}{}{}.
\newblock
{\BBOQ}\APACrefatitle {Meaningful questions: The acquisition of auxiliary
  inversion in a connectionist model of sentence production} {Meaningful
  questions: The acquisition of auxiliary inversion in a connectionist model of
  sentence production}.{\BBCQ}
\newblock
\APACjournalVolNumPages{Cognition}{166}{}{225--250}.
\PrintBackRefs{\CurrentBib}

\bibitem [\protect \citeauthoryear {%
Frank%
\ \BBA {} Mathis%
}{%
Frank%
\ \BBA {} Mathis%
}{%
{\protect \APACyear {2007}}%
}]{%
frank2007}
\APACinsertmetastar {%
frank2007}%
\begin{APACrefauthors}%
Frank, R.%
\BCBT {}\ \BBA {} Mathis, D.%
\end{APACrefauthors}%
\unskip\
\newblock
\APACrefYearMonthDay{2007}{}{}.
\newblock
{\BBOQ}\APACrefatitle {Transformational Networks} {Transformational
  networks}.{\BBCQ}
\newblock
\BIn{} \APACrefbtitle {Proceedings of the {Workshop on Psychocomputational
  Models of Human Language Acquisition}.} {Proceedings of the {Workshop on
  Psychocomputational Models of Human Language Acquisition}.}
\PrintBackRefs{\CurrentBib}

\bibitem [\protect \citeauthoryear {%
Hochreiter%
\ \BBA {} Schmidhuber%
}{%
Hochreiter%
\ \BBA {} Schmidhuber%
}{%
{\protect \APACyear {1997}}%
}]{%
hochreiter1997}
\APACinsertmetastar {%
hochreiter1997}%
\begin{APACrefauthors}%
Hochreiter, S.%
\BCBT {}\ \BBA {} Schmidhuber, J.%
\end{APACrefauthors}%
\unskip\
\newblock
\APACrefYearMonthDay{1997}{}{}.
\newblock
{\BBOQ}\APACrefatitle {Long short-term memory} {Long short-term memory}.{\BBCQ}
\newblock
\APACjournalVolNumPages{Neural Computation}{9}{8}{1735--1780}.
\PrintBackRefs{\CurrentBib}

\bibitem [\protect \citeauthoryear {%
Kam%
, Stoyneshka%
, Tornyova%
, Fodor%
\BCBL {}\ \BBA {} Sakas%
}{%
Kam%
\ \protect \BOthers {.}}{%
{\protect \APACyear {2008}}%
}]{%
kam2008}
\APACinsertmetastar {%
kam2008}%
\begin{APACrefauthors}%
Kam, X\BHBI N\BPBI C.%
, Stoyneshka, I.%
, Tornyova, L.%
, Fodor, J\BPBI D.%
\BCBL {}\ \BBA {} Sakas, W\BPBI G.%
\end{APACrefauthors}%
\unskip\
\newblock
\APACrefYearMonthDay{2008}{}{}.
\newblock
{\BBOQ}\APACrefatitle {Bigrams and the richness of the stimulus} {Bigrams and
  the richness of the stimulus}.{\BBCQ}
\newblock
\APACjournalVolNumPages{Cognitive Science}{32}{4}{771--787}.
\PrintBackRefs{\CurrentBib}

\bibitem [\protect \citeauthoryear {%
Lewis%
\ \BBA {} Elman%
}{%
Lewis%
\ \BBA {} Elman%
}{%
{\protect \APACyear {2001}}%
}]{%
lewis2001}
\APACinsertmetastar {%
lewis2001}%
\begin{APACrefauthors}%
Lewis, J\BPBI D.%
\BCBT {}\ \BBA {} Elman, J\BPBI L.%
\end{APACrefauthors}%
\unskip\
\newblock
\APACrefYearMonthDay{2001}{}{}.
\newblock
{\BBOQ}\APACrefatitle {Learnability and the statistical structure of language:
  Poverty of stimulus arguments revisited} {Learnability and the statistical
  structure of language: Poverty of stimulus arguments revisited}.{\BBCQ}
\newblock
\BIn{} \APACrefbtitle {Proceedings of {BUCLD}.} {Proceedings of {BUCLD}.}
\PrintBackRefs{\CurrentBib}

\bibitem [\protect \citeauthoryear {%
Linzen%
, Dupoux%
\BCBL {}\ \BBA {} Goldberg%
}{%
Linzen%
\ \protect \BOthers {.}}{%
{\protect \APACyear {2016}}%
}]{%
linzen2016assessing}
\APACinsertmetastar {%
linzen2016assessing}%
\begin{APACrefauthors}%
Linzen, T.%
, Dupoux, E.%
\BCBL {}\ \BBA {} Goldberg, Y.%
\end{APACrefauthors}%
\unskip\
\newblock
\APACrefYearMonthDay{2016}{}{}.
\newblock
{\BBOQ}\APACrefatitle {Assessing the ability of {LSTMs} to learn
  syntax-sensitive dependencies} {Assessing the ability of {LSTMs} to learn
  syntax-sensitive dependencies}.{\BBCQ}
\newblock
\APACjournalVolNumPages{Transactions of the Association for Computational
  Linguistics}{4}{}{521--535}.
\PrintBackRefs{\CurrentBib}

\bibitem [\protect \citeauthoryear {%
Marcus%
}{%
Marcus%
}{%
{\protect \APACyear {2018}}%
}]{%
marcus2018}
\APACinsertmetastar {%
marcus2018}%
\begin{APACrefauthors}%
Marcus, G.%
\end{APACrefauthors}%
\unskip\
\newblock
\APACrefYearMonthDay{2018}{}{}.
\newblock
{\BBOQ}\APACrefatitle {Innateness, {AlphaZero}, and Artificial Intelligence}
  {Innateness, {AlphaZero}, and artificial intelligence}.{\BBCQ}
\newblock
\APACjournalVolNumPages{arXiv preprint arXiv:1801.05667}{}{}{}.
\PrintBackRefs{\CurrentBib}

\bibitem [\protect \citeauthoryear {%
Mitchell%
}{%
Mitchell%
}{%
{\protect \APACyear {1980}}%
}]{%
mitchell1980}
\APACinsertmetastar {%
mitchell1980}%
\begin{APACrefauthors}%
Mitchell, T\BPBI M.%
\end{APACrefauthors}%
\unskip\
\newblock
\APACrefYearMonthDay{1980}{}{}.
\newblock
\APACrefbtitle {The need for biases in learning generalizations} {The need for
  biases in learning generalizations}\ \APACbVolEdTR{}{\BTR{}}.
\newblock
\APACaddressInstitution{}{Rutgers University}.
\PrintBackRefs{\CurrentBib}

\bibitem [\protect \citeauthoryear {%
Perfors%
, Tenenbaum%
\BCBL {}\ \BBA {} Regier%
}{%
Perfors%
\ \protect \BOthers {.}}{%
{\protect \APACyear {2011}}%
}]{%
perfors2011}
\APACinsertmetastar {%
perfors2011}%
\begin{APACrefauthors}%
Perfors, A.%
, Tenenbaum, J\BPBI B.%
\BCBL {}\ \BBA {} Regier, T.%
\end{APACrefauthors}%
\unskip\
\newblock
\APACrefYearMonthDay{2011}{}{}.
\newblock
{\BBOQ}\APACrefatitle {The learnability of abstract syntactic principles} {The
  learnability of abstract syntactic principles}.{\BBCQ}
\newblock
\APACjournalVolNumPages{Cognition}{118}{3}{306--338}.
\PrintBackRefs{\CurrentBib}

\bibitem [\protect \citeauthoryear {%
Pullum%
\ \BBA {} Scholz%
}{%
Pullum%
\ \BBA {} Scholz%
}{%
{\protect \APACyear {2002}}%
}]{%
pullum2002}
\APACinsertmetastar {%
pullum2002}%
\begin{APACrefauthors}%
Pullum, G\BPBI K.%
\BCBT {}\ \BBA {} Scholz, B\BPBI C.%
\end{APACrefauthors}%
\unskip\
\newblock
\APACrefYearMonthDay{2002}{}{}.
\newblock
{\BBOQ}\APACrefatitle {Empirical assessment of stimulus poverty arguments}
  {Empirical assessment of stimulus poverty arguments}.{\BBCQ}
\newblock
\APACjournalVolNumPages{The Linguistic Review}{18}{1-2}{9--50}.
\PrintBackRefs{\CurrentBib}

\bibitem [\protect \citeauthoryear {%
Srivastava%
, Hinton%
, Krizhevsky%
, Sutskever%
\BCBL {}\ \BBA {} Salakhutdinov%
}{%
Srivastava%
\ \protect \BOthers {.}}{%
{\protect \APACyear {2014}}%
}]{%
srivastava2014}
\APACinsertmetastar {%
srivastava2014}%
\begin{APACrefauthors}%
Srivastava, N.%
, Hinton, G.%
, Krizhevsky, A.%
, Sutskever, I.%
\BCBL {}\ \BBA {} Salakhutdinov, R.%
\end{APACrefauthors}%
\unskip\
\newblock
\APACrefYearMonthDay{2014}{}{}.
\newblock
{\BBOQ}\APACrefatitle {Dropout: A simple way to prevent neural networks from
  overfitting} {Dropout: A simple way to prevent neural networks from
  overfitting}.{\BBCQ}
\newblock
\APACjournalVolNumPages{The Journal of Machine Learning
  Research}{15}{1}{1929--1958}.
\PrintBackRefs{\CurrentBib}

\bibitem [\protect \citeauthoryear {%
Sutskever%
, Vinyals%
\BCBL {}\ \BBA {} Le%
}{%
Sutskever%
\ \protect \BOthers {.}}{%
{\protect \APACyear {2014}}%
}]{%
sutskever2014}
\APACinsertmetastar {%
sutskever2014}%
\begin{APACrefauthors}%
Sutskever, I.%
, Vinyals, O.%
\BCBL {}\ \BBA {} Le, Q\BPBI V.%
\end{APACrefauthors}%
\unskip\
\newblock
\APACrefYearMonthDay{2014}{}{}.
\newblock
{\BBOQ}\APACrefatitle {Sequence to sequence learning with neural networks}
  {Sequence to sequence learning with neural networks}.{\BBCQ}
\newblock
\BIn{} \APACrefbtitle {Proceedings of {NIPS}.} {Proceedings of {NIPS}.}
\PrintBackRefs{\CurrentBib}

\end{thebibliography}

\appendix

\section{Supplementary Materials}

\subsection{Details of the Grammar}

\begin{figure}[th]
\begin{center}

\vskip 0.12in
\fbox{\parbox{6cm}{
S $\rightarrow$ NP VP .

NP $\rightarrow$ Det N

NP $\rightarrow$ Det N PP

NP $\rightarrow$ Det N RC

VP $\rightarrow$ Aux V\_intrans

VP $\rightarrow$ Aux V\_trans NP

PP $\rightarrow$ P Det N

RC $\rightarrow$ Rel Aux V\_intrans

RC $\rightarrow$ Rel Det N Aux V\_intrans

RC $\rightarrow$ Rel Aux V\_trans Det N

Det $\rightarrow$ $\{$the $|$ some $|$ my $|$ your $|$ our $|$ her$\}$

\hangindent=0.7cm
{N $\rightarrow$ $\{$newt $|$ newts $|$ orangutan $|$ orangutans $|$ peacock $|$ peacocks $|$ quail $|$ quails $|$ raven $|$ ravens $|$ salamander $|$ salamanders $|$ tyrannosaurus $|$ tyrannosauruses $|$ unicorn $|$ unicorns $|$ vulture $|$ vultures $|$ walrus $|$ walruses $|$ xylophone $|$ xylophones $|$ yak $|$ yaks $|$ zebra $|$ zebras $\}$}

\hangindent=0.7cm
{V\_intrans $\rightarrow$ $\{$giggle $|$ smile $|$ sleep $|$ swim $|$ wait $|$ move $|$ change $|$ read $|$ eat $\}$}

\hangindent=0.7cm
{V\_trans $\rightarrow$ $\{$entertain $|$ amuse $|$ high\_five $|$ applaud $|$ confuse $|$ admire $|$ accept $|$ remember $|$ comfort $\}$}

\hangindent=0.7cm
{Aux $\rightarrow$ $\{$can $|$ will $|$ could $|$ would  $\}$}

\hangindent=0.7cm
{P $\rightarrow$ $\{$around $|$ near $|$ with $|$ upon $|$ by $|$ behind $|$ above $|$ below  $\}$}

\hangindent=0.7cm
{Rel $\rightarrow$ $\{$who $|$ that  $\}$}

}
}
\caption{Context-free grammar for the no-agreement language. 
The grammar contains 6 determiners (\textit{Det}), 26 nouns (\textit{N}), 9 intransitive verbs (\textit{V\_intrans}), 9 transitive verbs (\textit{V\_trans}), 4 auxiliaries (\textit{Aux}), 8 prepositions (\textit{P}), and 2 relativizers (\textit{Rel}).
} \label{fig:cfg}
\end{center}
\end{figure}

Figure \ref{fig:cfg} contains the context-free grammar used to generate the no-agreement language. 120,000 unique sentences were generated from this grammar as the training set, with each example randomly assigned either the identity task or the question formation task. If a sentence was assigned to the question formation task and contained a relative clause on the subject, it was not included in the training set. 

The agreement language was generated from a similar grammar but with the auxiliaries changed to \textit{do}, \textit{does}, \textit{don't}, and \textit{doesn't}. In addition, to ensure proper agreement, the grammar for the agreement language had separate rules for sentences with singular subjects and sentences with plural subjects, as well as separate rules for relative clauses with singular subjects and relative clauses with plural subjects.

\begin{figure*}[t]
\centering
\begin{subfigure}[t]{0.47\textwidth}
\centering
\begin{tabular} {llp{1.4cm}}
& Identity & Question formation \\ \hline
\textbf{Intransitive:} & & \\
~~~~~No modifiers & 0.012 & 0.012 \\
~~~~~PP on subject & 0.122 & 0.125 \\
~~~~~RC on subject & 0.121 & 0.000 \\
\textbf{Transitive:} &&\\
~~~~~No modifiers & 0.040 & 0.040 \\
~~~~~PP on subject & 0.041 & 0.041 \\
~~~~~PP on object & 0.041 & 0.041 \\
~~~~~RC on subject & 0.041 & 0.000 \\
~~~~~RC on object & 0.041 & 0.041 \\
~~~~~PP on subject, PP on object & 0.040 & 0.040 \\
~~~~~PP on subject, RC on object & 0.041 & 0.041 \\
~~~~~RC on subject, PP on object & 0.041 & 0.000 \\
~~~~~RC on subject, RC on object & 0.040 & 0.000 \\ \hline
\end{tabular}
\caption{No agreement language} \label{tab:freqsnoagr}
\end{subfigure}%
~\quad
\begin{subfigure}[t]{0.47\textwidth}
\centering
\begin{tabular} {llp{1.4cm}}
& Identity & Question formation \\ \hline
\textbf{Intransitive:} & & \\
~~~~~No modifiers & 0.005 & 0.005 \\
~~~~~PP on subject & 0.125 & 0.123 \\
~~~~~RC on subject & 0.120 & 0.000 \\
\textbf{Transitive:} &&\\
~~~~~No modifiers & 0.040 & 0.041 \\
~~~~~PP on subject & 0.042 & 0.041 \\
~~~~~PP on object & 0.042 & 0.041 \\
~~~~~RC on subject & 0.042 & 0.000 \\
~~~~~RC on object & 0.042 & 0.042 \\
~~~~~PP on subject, PP on object & 0.041 & 0.042 \\
~~~~~PP on subject, RC on object & 0.042 & 0.041 \\
~~~~~RC on subject, PP on object & 0.042 & 0.000 \\
~~~~~RC on subject, RC on object & 0.042 & 0.000 \\ \hline
\end{tabular}
\caption{Agreement language} \label{tab:freqsagr}
\end{subfigure}
\vskip 0.12in
\caption{Frequencies of sentence types in the two training sets. PP stands for \textit{prepositional phrase} and RC stands for \textit{relative clause}. Each line lists all modifiers in the sentences in question, so for example \textit{PP on object} excludes cases where there is also a modifier on the subject.} \label{fig:hyps}
\end{figure*}

Figure~\ref{tab:freqsnoagr} shows how frequent each sentence type was based on the types of modifiers present in the sentence and which noun phrases those modifiers were modifying. Figure~\ref{tab:freqsagr} shows the same statistics for the agreement language. In general, for a given left-hand side in the grammar in Figure~\ref{fig:cfg}, all rules with that left-hand side were equally probable; so, for example, one third of noun phrases were unmodified, one third were modified by a prepositional phrase, and one third were modified by a relative clause. The one exception to this generalization is that intransitive sentences with unmodified subjects were rare in both languages. This is because we did not allow any repeated items within or across data sets, and since there were relatively few possible intransitive sentences with unmodified subjects, this uniqueness constraint prevented the unmodified intransitive case from being as common as the modified cases. The no-agreement language has roughly twice as many intransitive sentences with unmodified subjects as the agreement language does because there are twice as many possible sentences of that type for the no-agreement language than the agreement language, but otherwise the two languages are essentially the same in the distributions of their constructions.

Neither language exhibited recursion. This is because relative clauses and prepositional phrases could only modify matrix noun phrases but not noun phrases within relative clauses or prepositional phrases. Thus, both languages contained a finite number of sentences, though this finite number is very large (greater than $10^{15}$), orders of magnitude larger than the number of sentences present in the training set (120,000).

\subsection{Details of the Architecture}

\begin{figure*}
\begin{center} 
\includegraphics[width=\textwidth]{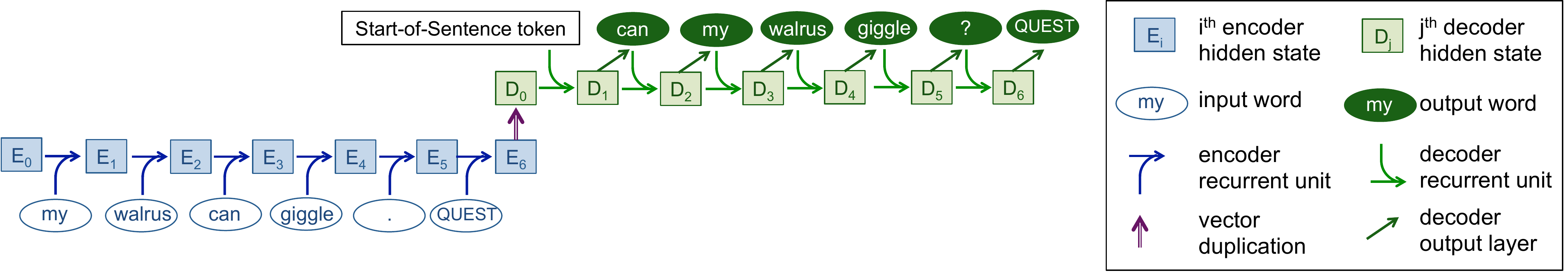}
\end{center}
\caption{Basic sequence-to-sequence neural network without attention.} \label{fig:seq2}
\end{figure*}

Figure \ref{fig:seq} (reproduced here as Figure \ref{fig:seq2}) depicts the basic sequence-to-sequence architecture underlying all of our experiments. Here we elaborate on the different components of this architecture. 

The network consists of two components, the \textbf{encoder} and the \textbf{decoder}, both of which are RNNs. The encoder's hidden state is initialized at E$_0$ as a 256-dimensional vector of all zeros. The network is then fed the first word of the input sentence, represented in a distributed manner as a 256-dimensional vector (i.e. an embedding) whose elements are learned during training. 
The encoder uses this distributed representation of the first word, along with the initial hidden state, to generate the next hidden state, E$_1$. The component that performs this hidden state update is called the encoder's \textbf{recurrent unit}.
Each subsequent word of the input sentence is then fed into the network, turned into its distributed representation learned by the network, and passed through the recurrent unit along with the previous hidden state to generate the next hidden state. 

Once all of the input words have been passed through the encoder, the final hidden state of the encoder is used as the initial hidden state of the decoder, D$_0$. This hidden state and a special start-of-sentence token (also represented by a 256-dimensional distributed representation that is learned during training) are passed as inputs to the decoder's recurrent unit, which outputs a new 256-dimensional vector as the next decoder hidden state, D$_1$. A copy of this new hidden state is also passed through a linear layer whose output is a vector with a length equal to the vocabulary size. The softmax function is then applied to this vector (so that its values sum to~1 and all fall between 0 and 1). Then, the element of this vector with the highest value is taken to correspond to the output word for that timestep; this correspondence is determined by a dictionary relating each index in the vector to a word in the vocabulary. For the next time step of decoding, this just-outputted word is converted to a distributed representation and is then taken as an input to the decoder's recurrent unit, along with the previous decoder hidden state, to generate the next decoder hidden state and the next output word. Once the outputted word is an end-of-sequence token (either \textit{IDENT} or \textit{QUEST}), decoding stops and the sequence of outputted words is taken as the output sentence. At all steps of this decoding process, whenever a distributed representation is used, dropout \cite{srivastava2014} with a proportion of 0.1 is applied to the vector, meaning that each of its values will with 10\% probability be turned to 0. This practice is meant to combat overfitting of the network's parameters.

\begin{figure}
\begin{center} 
\includegraphics[width=0.48\textwidth]{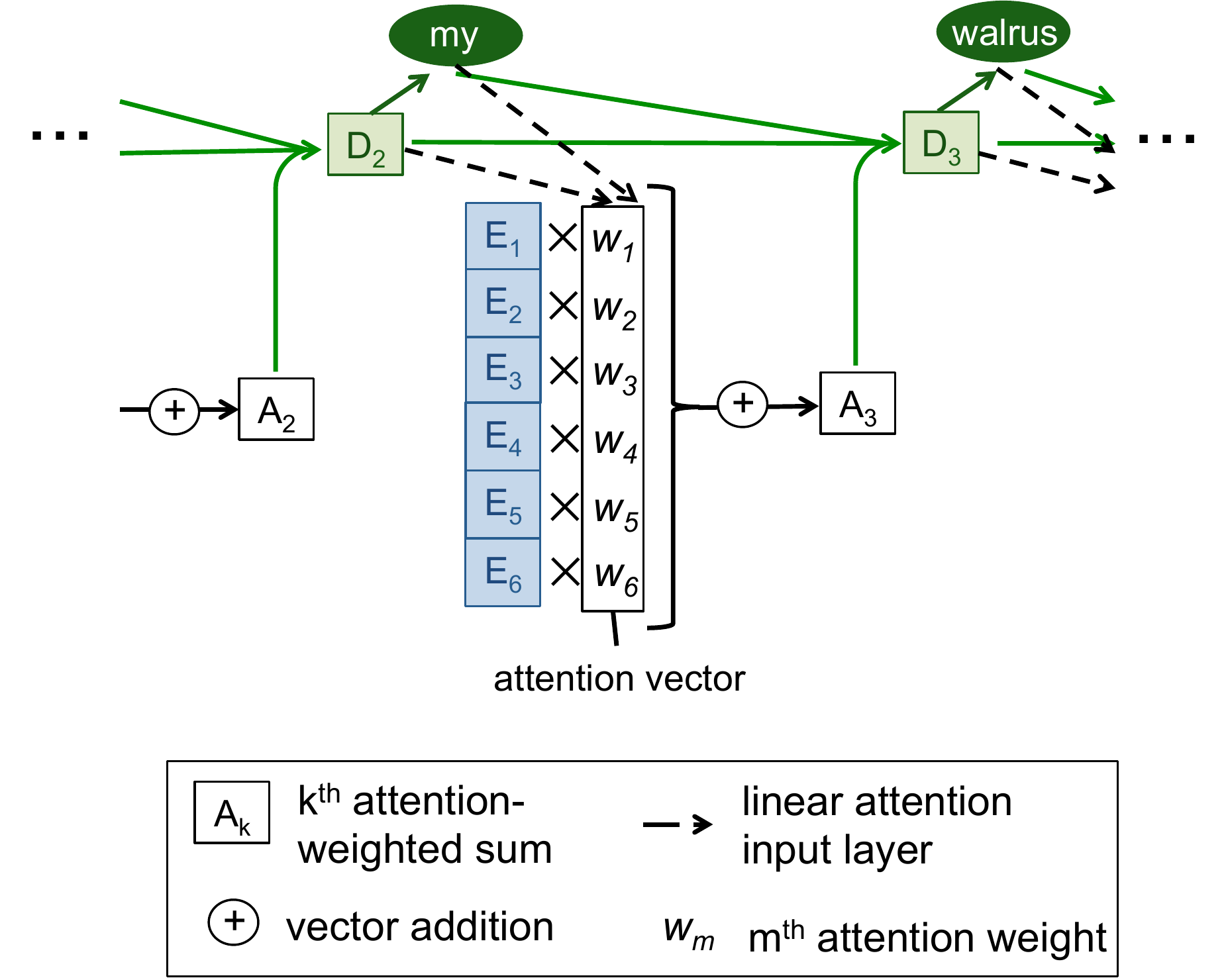}
\end{center}
\caption{The attention mechanism.} \label{fig:attn}
\end{figure}

There are two main ways in which we varied this basic architecture. First was usage of an attention mechanism, depicted in Figure \ref{fig:attn}, which is a modification to the decoder's recurrent unit. The attention mechanism adds a third input (which we refer to as the attention-weighted sum) to the decoder recurrent unit. This attention-weighted sum is determined as follows: First, the previous hidden state and the distributed representation of the previous output word  are passed through a linear layer whose output is a vector of length equal to the number of words in the input sentence. This vector is the vector of attention weights. Each of these weights is then multiplied by the hidden state of the encoder at the encoding time step equal to that weight's index. All of these products are then added together to give the attention-weighted sum, which is passed as an input to the decoder recurrent unit along with the previous output word and the previous hidden state.

Second, we also vary the structure of the recurrent unit used for the encoder and decoder. The three types of recurrent units we experiment with are simple recurrent networks (SRNs) \cite{elman1990}, gated recurrent units (GRUs) \cite{cho2014}, and long short-term memory (LSTM) units \cite{hochreiter1997}. For all three of these types of recurrent units, we use the default PyTorch implementations, which are described  in the next few paragraphs.

The SRN concatenates its inputs,  passes the result of the concatenation through a linear layer whose output consists of linear combinations of the elements of the input vector, and finally applies the hyperbolic tangent function to the result to create a vector whose values are mostly either very close to -1 or very close to +1. This hidden state update can be expressed with the following equation:

\begin{equation}
D_i = tanh(W * [D_{i-1}, w_{i-1}] + b)
\end{equation}

\noindent
where $D_i$ is the i$^{th}$ hidden state of the decoder, $w_i$ indicates the i$^{th}$ output word, $W$ is a matrix of learned weights, $b$ is a learned vector called the bias term, and $[v_1, v_2,...]$ indicates the concatenation of vectors $v_1$, $v_2$, .... If attention is used, this equation then becomes

\begin{equation}
D_i = tanh(W * [D_{i-1}, w_{i-1}, A_i] + b)
\end{equation}

\noindent
where $A_i$ is the i$^{th}$ attention-weighted sum.

The GRU adds several internal vectors called gates to the basic SRN structure. Specifically, these gates are called the reset gate $r_t$, the input gate $z_t$, and the new gate $n_t$, each of which has a corresponding matrix of weights ($W_r$ for $r_t$, $W_z$ for $z_t$, and two separate matrices $W_{nw}$ and $W_{nD}$ for $n_t$). The reset and input gates both take the previous hidden state and the previously outputted word (as a distributed representation) as inputs. The new gate also takes these two inputs as well as the reset gate as a third input. The next hidden state is then generated as the product of the input gate and the previous hidden state plus the product of one minus the input gate times the new gate. This can be thought of as the input gate determining which elements of the hidden state to preserve and which to change. The elements to be preserved are preserved through the term that is the product of the input gate times the previous hidden state, while the elements to be changed are determined through the term that is 
the product of one minus the input gate times the new gate; the new gate here determines what the updated values for these changed terms should be. Overall the GRU update can be expressed with the following equations ($\sigma$ indicates the sigmoid function):

\begin{equation}
r_t = \sigma(W_r * [D_{t-1}, w_{t-1}] + b_r)
\end{equation}
\begin{equation}
z_t = \sigma(W_z * [D_{t-1}, w_{t-1}] + b_z)
\end{equation}
\begin{equation}
n_t = tanh(W_{nw} * w_{t-1} + b_{nw} + r_t * W_{nD} * (D_{t-1} + b_{nD}))
\end{equation}
\begin{equation}
D_t = z_t * D_{t - 1} + (1 - z_t) * n_t
\end{equation}

Like the GRU, the LSTM also uses gates---specifically, the input gate $i_t$, forget gate $f_t$, cell gate $g_t$, and output gate $o_t$. Furthermore, while the other architectures all just use the hidden states as the memory of the network, the LSTM adds a second vector called the cell state $c_t$ that acts as another persistent state that is passed from time step to time step. These components interact according to the following equations to produce the next hidden state and cell state:

\begin{equation}
i_t = \sigma(W_i * [D_{t-1}, w_{t-1}] + b_i)
\end{equation}
\begin{equation}
f_t = \sigma(W_f * [D_{t-1}, w_{t-1}] + b_f)
\end{equation}
\begin{equation}
g_t = tanh(W_g * [D_{t-1}, w_{t-1}] + b_g)
\end{equation}
\begin{equation}
o_t = \sigma(W_o * [D_{t-1}, w_{t-1}] + b_o)
\end{equation}
\begin{equation}
c_t = f_t * c_{t-1} + i_t * g_t
\end{equation}
\begin{equation}
D_t = o_t * tanh(c_t)
\end{equation}

For each pair of an architecture and a language, we trained 100 networks with different random initializations, for a total of 1200 trained networks. The networks were trained using stochastic gradient descent with the negative log likelihood objective function for 30,000 batches with a batch size of 5 (meaning that some training examples were seen more than once),
a dropout rate of 0.1, and a learning rate of 0.01 (for the GRUs and LSTMs) or 0.001 (for the SRNs).
All networks used 256-dimensional hidden states and trained 256-dimensional vector representations of words. All parameter values were taken from a PyTorch tutorial on sequence-to-sequence networks,\footnote{\url{https://bit.ly/2I9WKBg}} except that the learning rate for SRNs was lowered because these networks did not converge with the default learning rate.

\subsection{Test and Generalization Accuracies}

Table \ref{tab:fullacc} gives the accuracies of various architectures on the test set and generalization set.

\begin{table}[!ht]
\begin{center} 
\begin{tabular}{lC{1cm}C{1cm}C{1cm}C{1cm}} 
    \toprule
\multirow{2}{*}{}    &  \multicolumn{2}{c}{Test set} & \multicolumn{2}{c}{Generalization set} \\ \cmidrule{2-5}
& Word match & POS match & Word match & POS match \\
    \midrule
SRN      & 0.001 & 0.811  & 0.000 & 0.000 \\
SRN + attn  & 0.942 & 0.999 & 0.010  & 0.023 \\
GRU       & 0.983 & 0.998 & 0.008 & 0.022 \\
GRU + attn        & 0.975 & 0.993  & 0.033 & 0.041 \\
LSTM & 0.999 & 1.000 & 0.046 & 0.069  \\
LSTM + attn & 0.998 & 1.000 & 0.133 & 0.185 \\
\bottomrule
\end{tabular}
\vskip 0.12in
\caption{Accuracy of each architecture with the agreement language. \textit{Word match} means getting the output exactly correct, while \textit{POS match} allows words to be replaced by other words of the same part of speech (POS). Each number in this table is an average across 100 networks.}
\label{tab:fullacc} 
\end{center} 
\end{table}

\subsection{Details of the Linear Classifiers}

Each linear classifier consisted of a single linear layer which took as its input a 256-dimensional vector (specifically, the encoding of a sentence) and outputted a vector of dimension equal to the number of possible values for the feature used as the basis of classification (4 for the main auxiliary, 28 for the fourth word, or 26 for the subject noun). For example, since there are four auxiliaries, the main auxiliary classifier had an output of dimensionality 4. The chance baseline for each task is thus $\frac{1}{n}$ where $n$ is the number of possible classes for that task. Each element in this output corresponded to a specific value for the feature being used as the basis for classification, and for a given input the element of the output with the highest value was taken as the classification for that input. 
The sentence encodings were randomly split into a training set (75\% of the encodings), a development test set (5\% of the encodings), and a test set (20\% of the encodings), none of which overlapped.
The weights of the classifier were trained on the training set using stochastic gradient descent, and training stopped when the cross entropy loss computed over the development test set ceased to improve. Classification accuracy was then determined based on the withheld test set. In addition to the information gleaned from the sentence encoding, the decoder may also access information about the input sentence through attention, but here we did not analyze the contribution of attention because it was not present in all architectures and because we wished to use an analysis method that could compare all six architectures.

\subsection{Examples of uncategorized errors}

Table \ref{tab:candy} categorized the 65\% of GRU with attention outputs that could be viewed as taking the input, inserting an auxiliary at its start, and deleting either zero or one auxiliary from within the input. Here we show some examples of the remaining 35\% of the outputs. Many of these outputs fall into certain common error types, including truncation of the output (\ref{ex:trunc}), repetition of words (\ref{ex:rep}), and turning a relative clause into a prepositional phrase (\ref{ex:reltopp}), but many outputs do not fall into one of those categories (\ref{ex:other1}, \ref{ex:other2}, \ref{ex:other3}).

\begin{exe}
\ex \label{ex:trunc}\textbf{Truncation of the output}\begin{xlist}
\ex \textit{Input:} my yak who our salamander does admire does read .
\ex \textit{Output:} does my yak who our salamander does admire ?
\end{xlist}
\ex \label{ex:rep}\textbf{Repetition of words}\begin{xlist}
\ex \textit{Input:} my walrus who doesn't sleep does sleep .
\ex \textit{Output:} does my walrus sleep sleep sleep ?
\end{xlist}
\ex \label{ex:reltopp}\textbf{Turning a relative clause into a prepositional phrase}\begin{xlist}
\ex \textit{Input:} my tyrannosauruses that the zebra does confuse do admire my walrus behind her orangutans .
\ex \textit{Output:} don't my tyrannosauruses behind the zebra confuse your my her walrus behind her orangutans ?
\end{xlist}
\ex \label{ex:other1}\textbf{Other}\begin{xlist}
\ex \textit{Input:} the newt who some quails do accept doesn't comfort the ravens that do move .
\ex \textit{Output:} do the newt who some quails doesn't accept doesn't comfort the ravens that do high\_five our yaks .
\end{xlist}
\ex \label{ex:other2}\textbf{Other}\begin{xlist}
\ex \textit{Input:} the yak who does applaud your newts doesn't entertain our walruses . 
\ex \textit{Output:} does the yak who the yak newts doesn't entertain our walruses ?
\end{xlist}
\ex \label{ex:other3}\textbf{Other}\begin{xlist}
\ex \textit{Input:} your vulture who does entertain our orangutan doesn't comfort my walruses above your newts .
\ex \textit{Output:} does your vulture entertain some peacock upon our ravens my walruses ?
\end{xlist}
\end{exe}

\end{document}